# Explicit and Implicit Models in Infrared and Visible Image Fusion

Zixuan Wang [1, 2]

1 School of Aeronautics and Astronautics, University of Electronic Science and Technology of China
Chengdu 611731, China
2 Aircraft Swarm Intelligent Sensing and Cooperative Control Key Laboratory of Sichuan Province
Chengdu 611731, China
zixuan_wang0716@163.com

Bin Sun [1, 2, *]

1 School of Aeronautics and Astronautics, University of Electronic Science and Technology of China
Chengdu 611731, China
2 Aircraft Swarm Intelligent Sensing and Cooperative Control Key Laboratory of Sichuan Province
Chengdu 611731, China
sunbinhust@uestc.edu.cn

*Abstract*—**Infrared and visible images, as multi-modal image pairs, show significant differences in the expression of the same scene. The image fusion task is faced with two problems: one is to maintain the unique features between different modalities, and the other is to maintain features at various levels like local and global features. This paper discusses the limitations of deep learning models in image fusion and the corresponding optimization strategies. Based on artificially designed structures and constraints, we divide models into explicit models, and implicit models that adaptively learn high-level features or can establish global pixel associations. Ten models for comparison experiments on 21 test sets were screened. The qualitative and quantitative results show that the implicit models have more comprehensive ability to learn image features. At the same time, the stability of them needs to be improved. Aiming at the advantages and limitations to be solved by existing algorithms, we discuss the main problems of multi-modal image fusion and future research directions.**

## I. Introduction

Infrared and visible images are multi-modal image pairs with different expressions obtained by different sensors shooting the same scene. Specifically, the infrared images contain the thermal radiation information of objects. Visible images containing texture details and gradient information are more in line with human visual effects. For example, the semantic and profile features of pedestrian are self-explanatory in infrared images (Fig. 1 (a)), but weak in visible images. Similarly, we can observe the object's texture details at high resolution in the visible image (Fig.1 (b)) rather than in the infrared image. Generally, the unique features of visible images refer to high-resolution and local features, which can be extracted well by simple networks. And the unique features of infrared images refer to high-level semantic and global contour features, which require more complex and deep networks to extract. These unique features of multi-modal images lead to information redundancy, complementarity, and crossover.

This work was sponsored by Sichuan Science and Technology Program under Grant 2020YFG0231.

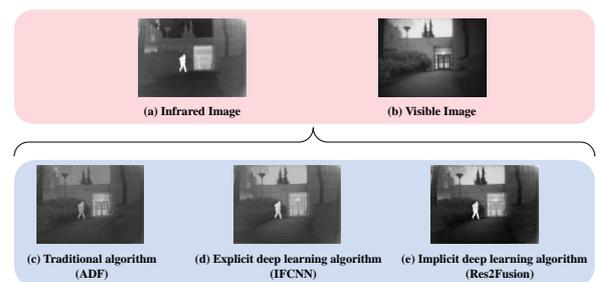

Fig.1. Infrared and visible image fusion research results in different stages. (a, b) are two source images, (c) shows a fused image based on the traditional algorithm, (d) shows a fused image based on an explicit neural network, and (e) shows a fused image based on an implicit neural network based on spatial and channel attention mechanism.

Recently, Y. Huang et al.[1] have shown multi-modality data can learn latent representations better than single-modality data in specific linear regression models. It is assumed that in a series of multi-modal learning tasks, information of different modalities is reasonably processed and fused. In this case, the processed data will show richer information than the unprocessed single-modal data. Researchers have applied visible and infrared image fusion to target recognition and detection, surveillance, and satellite remote sensing and other fields. They mainly used traditional algorithms to complete the fusion task at the beginning of the research. However, they are too weak to distinguish and extract special features. Furthermore, some advanced features cannot be processed well by traditional algorithms[2] (Fig.1. (c)). Deep learning methods with continuous improvement are gradually becoming the focus of research in image fusion tasks in recent years due to their solid ability to find the relationship between multi-modal data and learn feature representation. Generally, researchers design specific networks to extract target features and special optimization functions to improve network performance. Combined with these artificial constraints, the network usually trains in the desired direction and shows desired results. We call them explicitly neural networks that offer better results than traditional models. However, it comes with some drawbacks.

Firstly, there are limitations in feature extraction stage. The traditional convolutional neural network improves the expression ability of nonlinear features and performs well in extracting local detail features. Nonetheless, they extract higher-level semantic and global features by layering and deepening networks, causing problems like gradient disappearance or explosion and degrade the performance of networks. At the same time, information about the middle layer will be lost if only the characteristics of the last layer in the network are retained and output. It is necessary to optimize the network structure. The short connection in ResNet[3] mitigates the degradation problems like gradient disappearance or explosion caused by deepening the network. DenseNet[4] transfers more information from layer to layer with fewer parameters. The two networks and their combinations (such as RDN[5] and RXDNFuse[6]) have prevented network degradation. Other structures including U-Net[7], feature pyramid network[8], and nest connection architecture[9] retain more intermediate information effectively. These networks are pre-trained autoencoders or added as more concise modules to the encoder network's internal design.

Secondly, researchers rarely consider differences between modalities. In most cases, they apply the same encoder or the Siamese network to extract source images' features. As a result, they often extract the same features and ignore their differences. Some researchers consider using different networks to decouple features to deal with these problems. After disentangling features, the encoder is usually designed with a multi-stream structure and follows different fusion strategies. Feature decoupling has advantages in processing multimodal data and distinguishing modal differences theoretically. However, it is not easy to design an encoder network to extract target features with good interpretability. How much overlap is there in the characteristics of target separation? Can they completely reconstruct the information of the source image? These challenges need to be addressed soon for multi-modal data fusion.

These are the explicit deep learning models we mentioned, which are mostly in the form of encoder-decoder network structures. By changing network design, optimization direction, and training data, we can observe and control the influence of the model on the results. However, we want an ideal deep learning model with better utilization that can automatically learn high-level and global features from the dataset, which we call implicit models (Fig.1. (e)). It is not easy to accomplish with simple neural networks or artificial strategies but more complex networks.

Different from CNN artificially setting of the receptive field, the self-attention mechanism considers the global correlation between the whole image pixels to automatically find the target area. (In the fusion modules, spatial attention, channel attention, and their combination models are more commonly used to correlate spatial position and exchange channel information.) The transformer was proposed based on the self-attention mechanism, which integrates multi-head attention and adds residual connections to prevent further gradient explosion. The normalization operation effectively prevents the gradient from disappearing during calculation. Although the self-attention and transformer models improve parallelism, global correlation, and long-term dependencies, there are also significant restrictions on extracting local features. Generally, they combine with CNN to get multi-level features. Moreover, the amount of calculation is large since entire image pixels are considered during each analysis. The model is more flexible and requires large training datasets to achieve good results.

Besides, unlike other computer vision tasks, there are no ground truth images in image fusion. However, the generative models[10, 11] can recover the joint probability distribution and estimate the posterior probability distribution regardless of whether the original datasets are labeled or not. The popular generative model in image fusion is GAN consisted of two networks that need to be trained separately, called generator and discriminator. The model automatically discovers and learns high-level features through alternating training of two networks and generates new images with essential features. However, the two networks are difficult to balance and monitor during the training phase. We can't keep up with the quality of the generated images, which leads to the model not converging. The corresponding optimization introduces Wasserstein distance into the loss function and makes generator optimization more obvious.

This paper summarizes the limitations of deep learning models and the related optimization schemes (TABLE I). We divided these models into explicit and implicit models according to their characteristics and training methods. Furthermore, the qualitative and quantitative results of the 10 models illustrate the significant difference between the explicit and implicit models. Based on the advantages and disadvantages of these models, we analyzed the future research direction of multi-modal image fusion. In section II, we review specific explicit and implicit models applied in image fusion and classify them into several categories. In section III, we choose 10 state-of-the-art models and experiment on 21 image pairs. Then the qualitative and quantitative results in six metrics are analyzed. In section IV, we illustrate the conclusion and future research directions.

TABLE I. ANALYSIS OF DEEP LEARNING MODELS

| Deep Learning Algorithms | Model analysis | |
|---|---|---|
| | *Limitations* | *Optimization* |
| Explicit models | Gradient explosion/disappearance | Residual/short connection; pre-training |
| Explicit models | Weak global information | Self-attention mechanism; transformer models |
| Explicit models | Ignore differences in modalities | Disentangle features; information interaction |
| Explicit models | Lack ground-truth images | Generative models |
| Implicit models | Complex calculations | Dimensionality reduction for secondary variables |
| Implicit models | Weak local features | Combine with explicit neural networks |
| Implicit models | Poor stability | Optimization loss function (e.g., WGAN) |
| Implicit models | The training phase is difficult to monitor | Regularize network structure and training methods |

## II. EXPLICIT AND IMPLICIT MODELS REVIEW

### A. Explicit Models

Various encoder-decoder networks based on deep convolutional neural networks have been applied to infrared and visible image fusion in recent years. The basic structure includes encoders, fusion blocks, and decoders. The encoders designed for specific network structures are generally responsible for extracting target features. Then specific fusion strategies or modules are integrated into the networks to generate fused features. The decoders are in charge of limiting the size of a fused image of as the source images from the fused features.

H. Li et al.[12] divided source image pairs into basic parts and detailed content. The encoder network adopted the previously trained VGG-19 network to extract multi-layer features in detailed content. Similarly, J. Zhou et al.[13] design a model using the VGG-19 network as the backbone. On this basis, H. Li et al.[14] used the zero-phase component analysis method, which has the advantage of mapping features to a subspace, facilitating feature classification and reconstruction. The backbone of the encoder network was replaced by ResNet50 and ResNet101. Similarly, K. Ren et al.[15] combine DenseNet with the zero-phase component analysis method.

Y. Zhang et al.[16] proposed a unified model named IFCNN for different image fusion tasks. They used Siamese network in the encoder and added perceptual loss, making the fusion result contain more texture information and is in line with human visual effect. Siamese networks extracted the same type of features. L. Jian et al.[17] design a symmetrical network based on this structure. Specifically, they added a residual block to the encoder network to achieve optimal convergence in the training phase and fully use the previous features. H. Li et al.[18] introduce a network based on DeepFuse[19] and design two fusion strategies. Due to the addition of dense blocks, the encoder network fully considers the relationship between layers to extract features in the middle layers more completely. At the same time, regularization effect reduces over-fitting in dense connections. W. AN et al.[20] designed dense blocks in the encoder network. Residual blocks improved training stability, and the dense connection gave rise to the complete extraction of multi-scale features. Will their combination make the network have better performance? Y. Long et al.[6] introduce an end-to-end model using a variant of the residual network called ResNeXt[21]. Hafiz et al.[22] proposed a dual-stream network with dilated convolution instead of traditional convolution. They introduced the attention mechanism to the fusion layer to obtain global features and complementary features. H. Li et al.[23] integrated the "U-Net++" architecture into the network. NestFuse can save a large amount of feature information of the input image from a multi-scale perspective because the "U-Net++" network is more fully featured in the lower-level feature transformation. At the same time, the nest connections effectively solved the semantic gap caused by long skip connections.

In multimodal images fusion, the information of each source image is not exactly the same. If use the same module is used to extract the same type of features, it will lose the unique features of different modal images. What information should be retained and integrated? Researchers designed the dual-stream or even multi-stream encoder network based on these considerations. For example, aiming at the problem of the loss of high-level semantic features of images, Y. Fu et al.[24] introduced a dual-stream branch network to extract the detailed and semantic information respectively. H. Xu et al.[25] proposed a disentangled representation for the source image pairs. They designed multi-stream encoders to extract scene and attribute features, then fused them with respective rules. Considering the different content of source images, they introduced the pseudo-Siamese network (network structures are the same, but the weights are not the same exactly) in scene encoders. L. Ren et al.[26] proposed a network based on variational autoencoders. Residual blocks and symmetric skip connections were added to the network to avoid gradient disappearance and to improve the efficiency of training.

We classified the explicit models based on the "encoding-fusion-decoding" structure into four categories (Fig.2.): (a) The modal differences of source images are ignored. This type of network[14, 15, 18, 20, 27, 28] is generally designed with an encoder to extract features and a fusion rule to fuse features. (b) Encoders adopted the same network structure[16, 17], but the fusion rules are not specific to different encoders. (c) Unique features are extracted by encoders with different networks and rules[24, 25]. (d) Global features in multi-scale levels are extracted. Unlike the algorithms that design fusion rules or fusion modules, H. Xu et al.[29] introduced U2Fusion to generate fusion images by introducing densely connected networks.

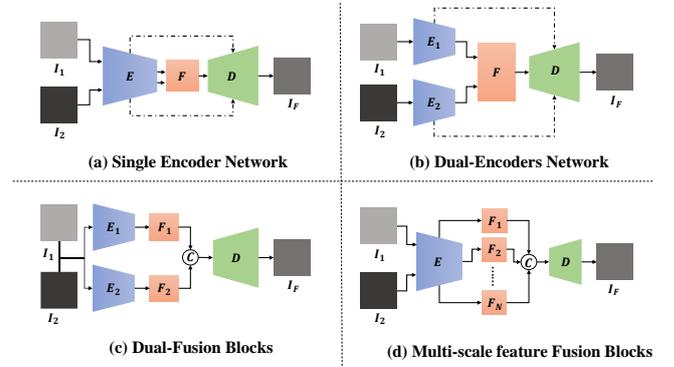

Fig.2. Classification of explicit models in IR-VI image fusion. The dashed line means that the decoder network in some literature considers the original feature mapping problem in the source image.

### B. Implicit Models

Recently, the self-attention mechanism has shown significant advantages in global feature extraction and has been applied to advanced feature tasks like semantic segmentation. In image fusion, spatial attention and channel attention are mainly employed to correlate spatial position and exchange channel information. For example, the fusion module of NestFuse consists of spatial and channel attention modules. H. Li et al.[30] proposed an end-to-end network RFN-Nest based on residual block to improve fusion strategy and make them learnable. J. Fu et al.[31] proposed a dual attention network to acquire long-range information in the channel and spatial dimensions. Z. Wang et al.[32] proposed Res2Fusion based on this non-local fusion module. Y. Wu et al. proposed a dual attention mechanism network composed of position and channel attention modules. They designed a perception-based fusion rule. The self-attention and transformer models have weak ability to extract local features, but are good at the global correlation and long-term dependencies. V. VS et al.[33] introduced the IFT model based on the nest connection structure of RFN-Nest. They introduced the transformer module in the fusion layer to strengthen long-range information fusion while retain the local feature. H. Zhao et al.[34] introduced DNDT that extracted local features with DenseNet and global features with dual-transformer networks.

The typical generative adversarial network consists of a generator and a discriminator. In the image fusion tasks, the generator network generates a pre-fused image. The discriminator is responsible for comparing the difference between the pre-fused and ground-truth images. In general, the ground-truth image is the source images. The scalar output of the discriminator shows the similarity between the pre-fused image and the ground-truth image. The major challenges of GAN are designing the generator and the discriminator network and optimizing the loss function to balance the training phase. The early application of generative adversarial networks in image fusion occurred in 2018 when J. Ma et al.[35] propose FusionGAN. Models are suitable for the same resolution images and show better performance in different resolutions images. They proposed Pan-GAN[36] to fuse low-resolution multi-spectral images and high-resolution panchromatic images. Meanwhile, DDcGAN[37] consisted of two discriminators was proposed for low-resolution infrared images and high-resolution visible images fusion. The content loss function includes the TV norm, which preserves the gradient change of the visible image in the fused image. J. Li et al.[38] proposed the D2WGAN model based on the dual discriminator network and Wasserstein distance.

The design and optimization of the generator networks primarily determine the quality of pre-fused images. For example, Y. Fu et al.[39] added the DenseNet structure to the generator network to extract the features of the source image more comprehensively and deeply. Q. Li et al.[40] introduced the dual generator structure into the network RCGAN. One of the generators was used to extract thermal radiation information, and the other was to extract gradient information. The DSAGAN network proposed by J. Fu et al.[41] separated the source image pairs into the dual-stream network branch of the generator to extract multi-scale features and then used the attention mechanism to perform "feature enhancement" and "feature integration" respectively. Y. Cui et al.[42] proposed a channel attention network to deal with the drawback that the convolution filter in the non-attention mechanism has the same effect on each channel. A. Fang et al.[43] regarded the image fusion tasks as multi-task learning. They introduced a network based on the subjective visual attention mechanism. J. Li et al.[44] introduced the attention mechanism into the discriminator. They designed a multi-scale attention mechanism network in the generator to obtain the multi-scale attention maps of the source images. The model pays more attention to the local areas with prominent characteristics than to the entire image content in the fusion phase. In addition, they still retained the dual discriminator design. On this basis, they proposed MgAN-Fuse[45], integrating the attention mechanism fusion module into the multi-scale layers of the encoder. S. Yi et al.[46] design DFPGAN, obtaining the contrast images calculated by subtracting between the source image pairs. In addition, they designed the dual self-attention module and put it in the last layer of the convolution block to refine the feature map and improve the fusion performance.

We classified the algorithms mentioned in Fig. 3. Several unnamed models were not included due to organizational difficulties. The number of the multi-stream networks is more than the single-stream networks in explicit and implicit models. Considering the modal differences between the source image pairs, the multi-stream network can retain most of the features so that the fused images are closer to our ideal results.

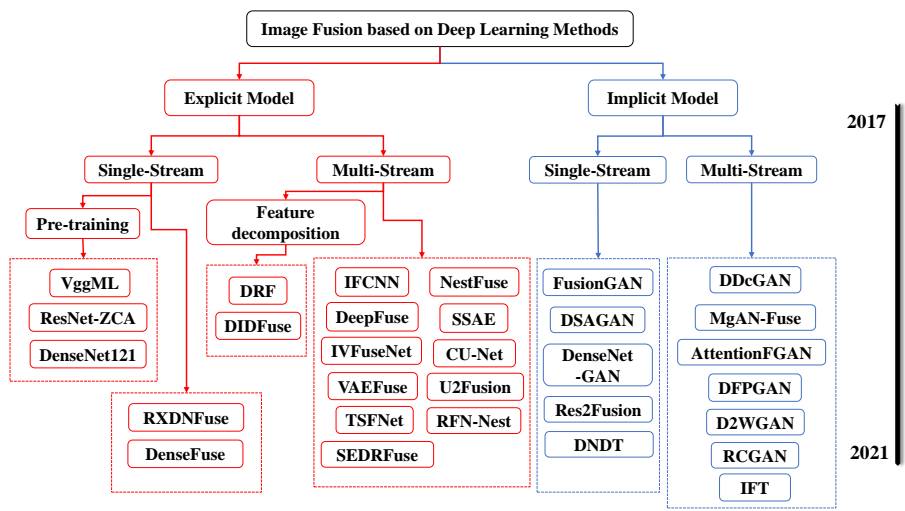

Fig.3. State-of-the-art deep learning algorithms of infrared and visible image fusion in recent years. The left part in red shows the explicit models, and the right in blue offers the implicit models.

## III. EXPERIMENT

In this section, we chose 21 pairs of images in the TNO dataset and ten publicly available state-of-the-art models based on deep learning. These models consist of explicit models (IFCNN, U2Fusion, Dual-add, ResNet-ZCA), combination with explicit network and implicit fusion strategy (NestFuse, IFT, SEDRFuse, Res2Fusion) and implicit models (DDcGAN, RCGAN). We considered that if there was a significant difference in performance between the explicit and implicit models, and if the implicit strategy was more helpful for extracting features comprehensively. The qualitative results are shown in Fig.4. We analyzed the objective evaluation with the benchmark named VIFB[47], including information theory-based metrics (CE, EN, MI, PSNR), structural similarity-based metrics (SSIM, RMSE), image feature-based metrics (AG, EI, SD, SF, $Q^{AB/F}$), and human perception inspired metrics ($Q_{CB}$, $Q_{CV}$). Here we select six metrics (CE, PSNR, RMSE, SD, $Q^{AB/F}$ and $Q_{CV}$) from them. CE, PSNR, RMSE, SD, $Q^{AB/F}$ and $Q_{CV}$ represent cross-entropy, peak signal-to-noise ratio, root mean squared error, standard deviation, gradient-based fusion performance, and Chen-Varshney metric. The quantitative results are shown in TABLE II. The best value in each metric is marked red, the second-best value in each metric in blue, and the third-best value in each metric in green.

The subjective evaluation was performed on the GeForce RTX 2080TI and 256G of CPU memory. In addition, the objective evaluation was implemented on MATLAB R2021a. The source image pairs and their subjective fusion results are shown in Appendix A.

Firstly, a horizontal analysis of each algorithm is shown in TABLE I. For IFCNN (Fig.4. (a, k)), we chose the elementwise-max fusion strategy for testing. Due to the perceptual loss, the fusion result produces more texture information and achieves the best in $Q^{AB/F}$. U2Fusion (Fig.4. (b, l)) with two similarity constraints to the loss function, has achieved third-best in RMSE and PSNR. It can be seen from the subjective results that U2Fusion keeps the texture detail information, and the brightness is balanced between the two source images. Dual-add (Fig.4. (c, m)) is the best in PSNR and RMSE. ResNet-ZCA (Fig.4. (d, n)) is the best in RMSE, and performs well in PSNR. NestFuse (Fig.4. (e, o)) retains multi-scale depth feature information because of the nested connection structure. It gets the second-best performance in $Q_{CV}$. In the implicit models, Res2Fusion (Fig.4. (g, q)) reaches the best in $Q^{AB/F}$ and $Q_{CV}$. DDcGAN (Fig.4. (i, s)) reaches the best in CE and SD, but not perform well in PSNR, $Q^{AB/F}$ or $Q_{CV}$. RCGAN (Fig.4. (j, t)) performs well in CE.

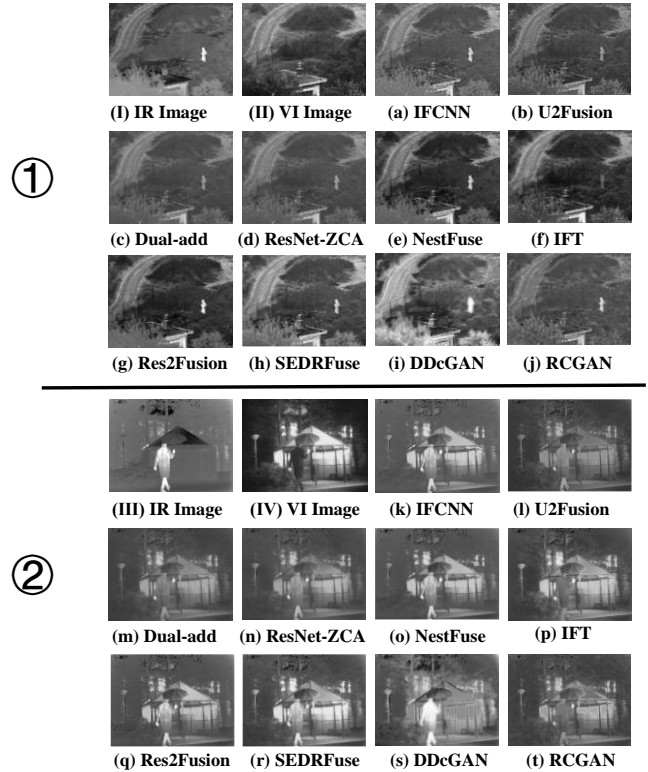

Fig.4. Qualitative comparison of 10 state-of-the-art deep learning algorithms on two typical Infrared and Visible image pairs in the TNO dataset.

TABLE II. THE AVERAGE EVALUATION METRIC VALUES OF 10 MODELS ON THE DATASET. (THE BEST VALUE IS MARKED IN RED, THE SECOND-BEST IN BLUE, AND THE THIRD-BEST IN GREEN)

| Models | Evaluation Metrics | | | | | |
|---|---|---|---|---|---|---|
| | *Information theory-based* | | *Structural similarity-based* | *Image feature-based* | | *Human perception inspired* |
| | **CE** | **PSNR** | **RMSE** | **SD** | **$Q^{AB/F}$** | **$Q_{CV}$** |
| IFCNN[16] | 1.608 | 59.044 | 0.084 | 31.400 | 0.504 | 312.294 |
| U2Fusion[29] | 1.418 | 59.376 | 0.078 | 22.472 | 0.427 | 618.592 |
| Dual-add[24] | 1.608 | 59.467 | 0.077 | 22.120 | 0.338 | 514.260 |
| ResNet-ZCA[14] | 1.488 | 59.456 | 0.077 | 23.742 | 0.374 | 447.528 |
| NestFuse[23] | 1.642 | 58.722 | 0.093 | 39.099 | 0.489 | 300.400 |
| IFT[33] | 1.422 | 55.707 | 0.179 | 38.992 | 0.467 | 510.315 |
| Res2Fusion[32] | 1.476 | 55.642 | 0.181 | 40.516 | 0.504 | 285.268 |
| SEDRFuse[17] | 1.291 | 58.863 | 0.089 | 38.295 | 0.466 | 447.978 |
| DDcGAN[37] | 0.951 | 54.889 | 0.110 | 45.816 | 0.371 | 890.635 |
| RCGAN[40] | 1.188 | 56.367 | 0.079 | 26.389 | 0.465 | 473.094 |

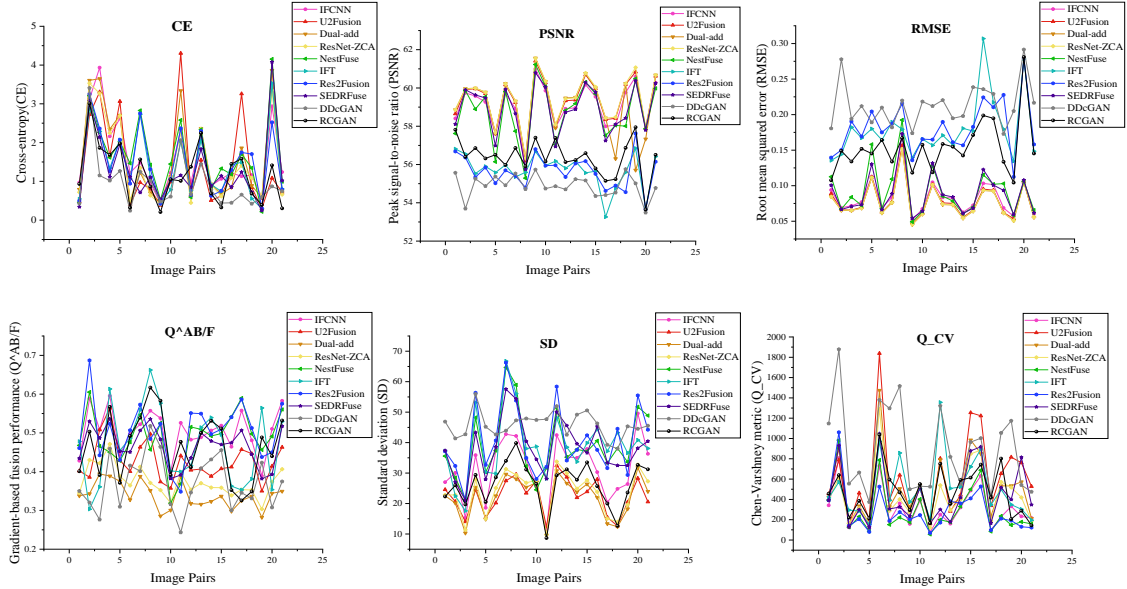

Fig.5. Quantitative comparisons of six metics of the ten state-of-the-art deep learning algorithms on 21 image pairs in the TNO dataset. The explicit models are with bright colors and implicit models with dark colors.

Secondly, we visualized and analyze the performance of ten algorithms on the same metric (Fig. 5.). We represent explicit models with bright colors and implicit models with dark colors. The explicit models like U2Fusion, dual-add and ResNet-ZCA are outperform the implicit models in PSNR and RMSE because of their specific designing or pre-training network. The implicit models significantly outperform the explicit models in CE, SD and $Q^{AB/F}$. Specially, the models combine with explicit network and implicit fusion strategy like NestFuse, IFT (Fig.4. (f, p)), SEDRFuse (Fig.4. (h, r)) and Res2Fusion are shown better performance than explicit models in SD, indicating that the self-attention and transformer modules are helpful to maintain multi-scale features. But they are not as clear as IFCNN and U2Fusion for some texture details such as the corners of buildings (Fig.4.). The combination of explicit and implicit models will be more comprehensive in feature extraction and learning than one of them. However, the implicit models are inferior to explicit models in PSNR and RMSE, illustrating their poor stability and need to be optimized in the future. There is no distinct difference between the explicit and implicit models in $Q_{CV}$.

## IV. CONCLUSION

This paper discusses the development of deep learning algorithms in infrared and visible image fusion in recent years. Firstly, we pointed out two essential factors that need to considered, including the different levels of local and global features in the single modality image, and the unique features of multi-modal images that cause modality differences. Meanwhile, we summarize the deep learning model's limitations and the corresponding optimization. Based on these characteristics, we classify the models into explicit and implicit models and give reasons.

In explicit models, the network structures, fusion strategies, and optimization functions are designed artificially. Based on the traditional convolutional neural networks, explicit models show robust ability of extracting local features. Nevertheless, the utilization of models is not high because they rely on artificial constraints. And the learning ability of global features is also poor. Generally, the global features are extracted by increasing the network depth, resulting in network degradation. In contrast, the implicit model has solid adaptive learning ability, which can establish long-term dependencies for the pixels of the entire image, thereby the global feature extraction and learning ability of the model is strong. It means that a large amount of data is required to train the model.

Secondly, we selected ten models from the state-of-the-art models and evaluated them on a unified test set and six objective evaluation metrics. It was found that the implicit models are superior to the explicit model in the image feature-based metrics. Nevertheless, it does not perform well in PSNR and structural similarity-based metrics. Implicit models improve the extraction ability of global features to a certain extent, while its stability is not high because of its robust adaptability and flexibility.

Finally, in the fusion task of multi-modal images, we summarize two critical aspects. On the one hand, we would better consider the difference of modalities. On the other hand, the differences between various levels of features within the modality and comprehensively extract local and global features should also be considered to ensure maximum retention of information. In the future research, we put forward suggestions to combine explicit models with implicit models to extract local and global features comprehensively, and to improve the stability of the implicit models.

ACKNOWLEDGMENT

This work was sponsored by Sichuan Science and Technology Program under Grant 2020YFG0231.